\documentclass{lmcs}
\pdfoutput=1
\usepackage[english]{babel}

\usepackage{graphicx}
\usepackage{booktabs}
\usepackage{multirow}

\usepackage[letterpaper,top=2cm,bottom=2cm,left=3cm,right=3cm,marginparwidth=1.75cm]{geometry}

\usepackage{amsmath}
\usepackage{graphicx}
\usepackage[colorlinks=true, allcolors=blue]{hyperref}

\title[Adaptive Spatio-Temporal Graphs for Weather Forecasting]{Adaptive Spatio-Temporal Graphs with Self-Supervised Pretraining for Multi-Horizon Weather Forecasting}
\author[Yao Liu]{Yao Liu\lmcsorcid{0009-0002-9180-4380}}[a]
\address[a]{School of Computer Science, Xiangtan University, Xiangtan, 411105, China}
\email[a]{202221632985@smail.xtu.edu.cn}

\begin{document}
\maketitle

\begin{abstract}
Accurate and robust weather forecasting remains a fundamental challenge due to the inherent spatio-temporal complexity of atmospheric systems. In this paper, we propose a novel self-supervised learning framework that leverages spatio-temporal structures to improve multi-variable weather prediction. The model integrates a graph neural network (GNN) for spatial reasoning, a self-supervised pretraining scheme for representation learning, and a spatio-temporal adaptation mechanism to enhance generalization across varying forecasting horizons. Extensive experiments on both ERA5 and MERRA-2 reanalysis datasets demonstrate that our approach achieves superior performance compared to traditional numerical weather prediction (NWP) models and recent deep learning methods. Quantitative evaluations and visual analyses in Beijing and Shanghai confirm the model’s capability to capture fine-grained meteorological patterns. The proposed framework provides a scalable and label-efficient solution for future data-driven weather forecasting systems.
\end{abstract}

\renewcommand{\thefootnote}{\fnsymbol{footnote}}
\setcounter{footnote}{-1}


\section{Introduction}

Accurate and timely weather forecasting is crucial for various societal functions, such as agriculture, transportation, emergency response, and urban planning. The inherently chaotic nature of weather systems, driven by complex, non-linear interactions across spatial and temporal scales, poses significant challenges for high-resolution forecasts. This is especially true in urban areas where localized phenomena such as heat islands and coastal effects further complicate predictions.

Traditional Numerical Weather Prediction (NWP) models, such as the European Centre for Medium-Range Weather Forecasts (ECMWF) and the Global Forecast System (GFS), have long been the backbone of weather forecasting. These models rely on solving the physical laws of atmospheric dynamics and thermodynamics through complex numerical simulations. Over time, NWP systems have seen substantial improvements in resolution and ensemble prediction techniques~\cite{eyre2020assimilation,zhang2010beating}. Despite these advancements, NWP models remain computationally intensive, which often limits their ability to provide real-time forecasts, especially in regions with complex terrains like urban and coastal areas~\cite{guichard2017short,mwanthi2024land}. Moreover, challenges such as integrating heterogeneous observations and accurately parameterizing sub-grid-scale phenomena persist and continue to be areas of active research~\cite{leutwyler2017evaluation}.

In parallel with NWP, statistical methods, including ARIMA and SARIMA models, have been applied for short-term weather forecasts by utilizing historical time series data~\cite{shivhare2019arima,rahman2013comparative}. These models, while effective in capturing linear trends, often struggle with the nonlinear and high-dimensional nature of atmospheric data. Hybrid approaches that combine statistical models with neural networks have been proposed, yet they typically depend on strong assumptions and handcrafted features.

The advent of deep learning (DL) has opened new avenues for modeling the complex spatio-temporal dynamics inherent in weather systems. Unlike NWP, DL methods are inherently data-driven, leveraging neural architectures to extract patterns from large-scale meteorological datasets. Convolutional Neural Networks (CNNs) have shown strong capabilities in spatial feature extraction, especially in satellite image analysis~\cite{kareem2021evaluation,fan2022using}. Recurrent Neural Networks (RNNs) and their advanced variants like LSTM and GRU are widely used for modeling temporal sequences~\cite{salman2015weather,devi2024ai}. Hybrid models, such as ConvLSTM~\cite{xingjian2015convolutional}, integrate convolutional and recurrent layers to forecast dynamic weather fields, performing well in tasks like nowcasting.

Further advancements include 3D CNNs~\cite{do2020stconvs2s}, which model spatio-temporal evolution of atmospheric variables, and transformer-based models like FourCastNet~\cite{pathak2022fourcastnet}, which utilize global attention mechanisms for large-scale weather prediction. Recent explorations into generative models, such as VAEs and GANs, have aimed to model uncertainty and produce diverse plausible future scenarios~\cite{vuyyuru2021novel, xiao2025vae, choi2024advancing}.

Despite these innovations, existing DL models face significant challenges. They often require vast amounts of labeled data for training, which can be difficult to obtain, and they may generalize poorly across varying regions or time scales. Additionally, these models frequently treat spatial and temporal dependencies separately, leading to difficulties in capturing the full complexity of evolving atmospheric systems. The limited incorporation of domain knowledge, such as physical conservation laws, also compromises both interpretability and robustness.

To tackle these challenges, we propose a spatio-temporal self-supervised learning framework for robust weather forecasting. Inspired by recent progress in generative models and hybrid spatio-temporal architectures~\cite{vuyyuru2021novel,xiao2025vae}, our method is tailored to learn discriminative and stable representations from reanalysis data without requiring labeled supervision. Key contributions of this work include:

\begin{itemize}
\item We introduce a self-supervised learning framework that exploits spatio-temporal dynamics to generate training signals without the need for labeled data. This addresses the challenge of requiring extensive labeled supervision and enhances the model's adaptability to diverse data sources.
\item We develop an adaptive mechanism that dynamically adjusts to short- and long-term forecast horizons and regional variations, improving the model's generalization capabilities across different geographic and temporal contexts.
\item We incorporate a graph neural network (GNN) module to effectively capture spatial dependencies across geographical regions, thereby enhancing forecast accuracy through unified modeling of spatio-temporal phenomena.
\end{itemize}

\section{Related Work}

\subsection{Traditional Numerical and Statistical Methods}

Traditional weather forecasting has long relied on Numerical Weather Prediction (NWP) models, which are grounded in the physical laws of atmospheric dynamics and thermodynamics. These models, such as the European Centre for Medium-Range Weather Forecasts (ECMWF)~\cite{haiden2017evaluation} and the Global Forecast System (GFS)~\cite{yue2022evaluation}, numerically solve partial differential equations using initial observations from satellite, radar, and ground stations. Over the past decades, NWP systems have seen major improvements in resolution, data assimilation, and ensemble prediction techniques~\cite{eyre2020assimilation,zhang2010beating,eyre2022assimilation}. High-resolution models operating at kilometer-scale grids offer greater detail in resolving small-scale processes like convection and turbulence~\cite{leutwyler2017evaluation}, while ensemble forecasting enables probabilistic forecasting, particularly valuable for extreme events such as hurricanes or heat waves.

Despite their strengths, NWP models remain computationally expensive and often struggle to deliver real-time or high-frequency forecasts in complex terrain like urban or coastal environments. Challenges such as parameterizing sub-grid-scale phenomena (e.g., cloud microphysics, land-atmosphere interactions) and integrating heterogeneous observations into initialization processes remain open research problems~\cite{guichard2017short, mwanthi2024land}. Furthermore, the inherent sensitivity to initial conditions makes long-term predictions particularly prone to uncertainty.

In parallel with NWP, statistical methods such as ARIMA, SARIMA, and multiple regression have been applied for short-term forecasts using historical time series~\cite{tektacs2010weather,shivhare2019arima,rahman2013comparative}. These models are effective in capturing linear relationships but often fall short when dealing with the nonlinear, high-dimensional nature of atmospheric data. Hybrid approaches combining linear models with neural networks have been proposed to enhance prediction accuracy, but they still rely on strong assumptions and handcrafted features.

\subsection{Deep Learning for Weather Forecasting}

In recent years, the rise of deep learning (DL) has provided new possibilities for modeling complex spatio-temporal dynamics in weather systems. Unlike NWP, DL methods are purely data-driven, relying on neural architectures to learn patterns from large-scale meteorological datasets. Convolutional Neural Networks (CNNs) have demonstrated strong capability in spatial feature extraction, particularly in satellite image analysis~\cite{kareem2021evaluation,fan2022using}, while Recurrent Neural Networks (RNNs) and their variants like LSTM and GRU have been widely used for temporal sequence modeling~\cite{salman2015weather,devi2024ai}.

Hybrid spatio-temporal models, such as ConvLSTM~\cite{xingjian2015convolutional}, combine convolutional and recurrent layers to forecast dynamic weather fields such as precipitation. These models have shown strong performance in short-range forecasting tasks like nowcasting. Further innovations include 3D CNNs~\cite{do2020stconvs2s}, which directly model the spatio-temporal evolution of atmospheric variables, and transformer-based models like FourCastNet~\cite{pathak2022fourcastnet}, which apply global attention mechanisms over latitude-longitude grids for large-scale weather prediction.
Recent work has also explored generative models, such as Variational Autoencoders (VAEs) and Generative Adversarial Networks (GANs), to model uncertainty and produce diverse plausible future scenarios~\cite{vuyyuru2021novel,xiao2025vae,choi2024advancing,hsieh2024developing,liu2023gan}. 

Despite these advances, deep learning methods still face several limitations. Many models require extensive labeled training data and are limited in their ability to generalize across regions or time scales. Moreover, most existing approaches model spatial and temporal dependencies independently, making it difficult to capture the full complexity of evolving atmospheric systems. There is also limited incorporation of domain knowledge, such as physical constraints or conservation laws, which can reduce interpretability and robustness.
In light of these limitations, there is a growing need for unified forecasting frameworks that can integrate spatio-temporal reasoning, self-supervised learning, and domain adaptability. Our work aims to address these challenges by proposing a novel deep learning architecture that jointly models spatial dependencies via Graph Neural Networks, learns from unlabeled data through contrastive objectives, and dynamically adapts across forecasting horizons through a spatio-temporal weighting mechanism.

\section{Methodology}

\subsection{Model Overview}

\begin{figure*}[htpb]
    \centering
    \includegraphics[width=0.8\linewidth]{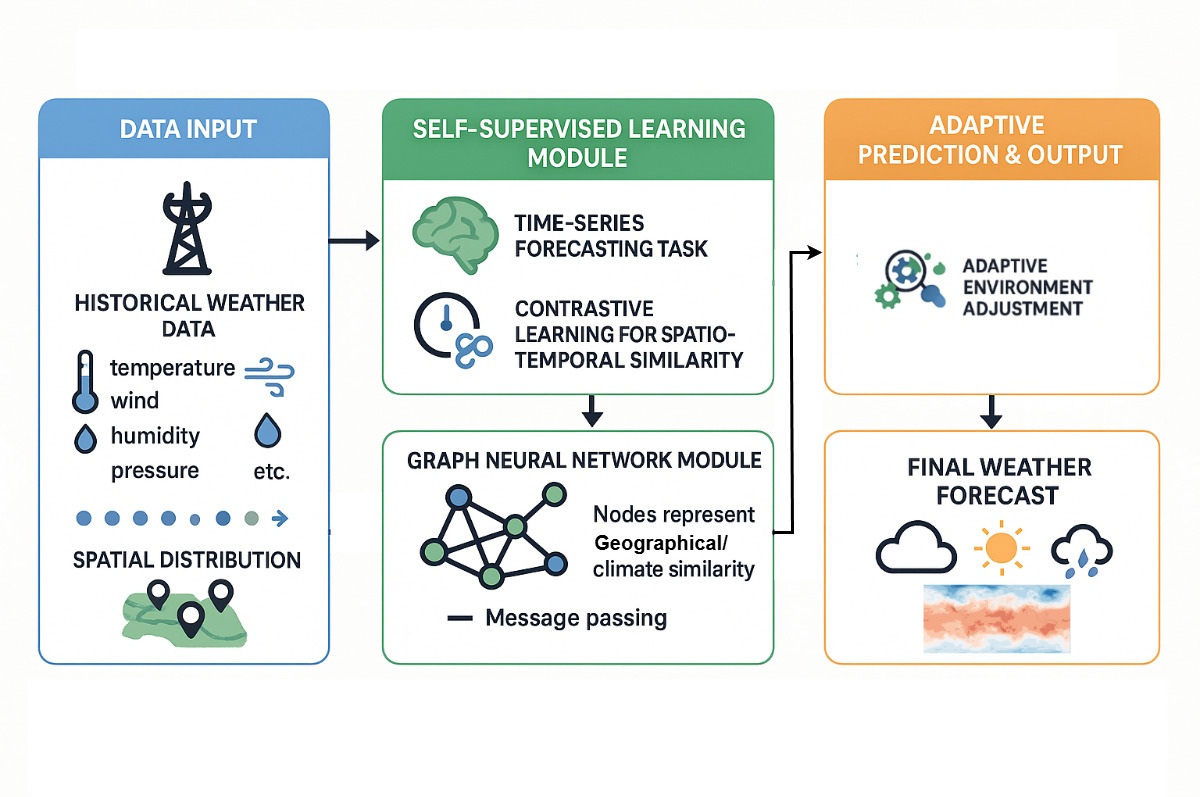}
    \caption{Overview of the proposed spatio-temporal self-supervised learning model for robust weather forecasting}
    \label{fig:model_overview}
\end{figure*}

The proposed Spatio-temporal Self-supervised Learning for Robust Weather Forecasting model integrates multiple components to enhance the accuracy and robustness of weather forecasting. As shown in Figure \ref{fig:model_overview}, the model consists of several key modules that work together in a sequential and adaptive manner to provide accurate and dynamic weather predictions.

As illustrated in Figure \ref{fig:model_overview}, the model begins with the Data Input stage, where historical weather data, including parameters such as temperature, wind speed, pressure, and humidity, are fed into the system. This data serves as the foundation for the following modules.
The Self-supervised Learning Framework generates its own forecasting targets from the input data, eliminating the need for labeled datasets. This framework uses temporal and spatial dependencies to predict future weather conditions. The model is trained using contrastive learning, which helps distinguish between different weather patterns and improves prediction accuracy.
Next, the Spatio-temporal Adaptation Mechanism dynamically adjusts the model’s learning strategy. For short-term predictions, the model places more emphasis on recent data, while for long-term predictions, it accounts for broader trends. This adjustment ensures the model performs well over various time horizons.
The Graph Neural Network (GNN) module captures the spatial dependencies between different geographical regions, allowing the model to understand how weather patterns in one area influence nearby regions. By propagating information across these regions, the GNN enhances the model’s ability to make accurate predictions on a larger scale.
Finally, the Adaptive Prediction and Output Generation module ensures that the model adapts continuously as new weather data becomes available. This enables the model to refine its forecasts in real-time, ensuring the predictions remain relevant and accurate as weather conditions change.

\subsection{Self-supervised Learning Framework}
The self-supervised learning framework is integral to the proposed model, enabling it to generate predictive targets directly from raw weather data without relying on manually labeled datasets. The primary innovation in this framework is the spatio-temporal target generation mechanism, where the model generates its own target predictions based on the temporal and spatial dependencies within the data, avoiding the need for predefined labels.

The spatio-temporal prediction task is defined as follows:
\begin{equation}
    \mathcal{L}(\mathcal{T}) = \sum_{t \in T} \mathbb{I}(y_t \sim \hat{y}_t) \cdot \left( \mathbf{f}_t - \hat{\mathbf{f}}_t \right)^2
\end{equation}
where \( \mathcal{L}(\mathcal{T}) \) represents the loss function over time \(T\), \(y_t\) is the actual weather observation at time \(t\), \( \hat{y}_t \) is the self-generated target prediction at time \(t\), and \( \mathbf{f}_t \) and \( \hat{\mathbf{f}}_t \) are the feature vectors representing the true and predicted weather conditions at time \(t\), respectively.

The model aims to minimize the difference between the predicted and actual weather feature vectors over time. The learning process is enhanced by contrastive learning, which helps the model distinguish between similar and dissimilar weather patterns across both time and space, improving the model’s ability to generalize and predict future states.

To further refine the learning, we introduce a contrastive loss:
\begin{equation}
    \mathcal{L}_{\text{contrastive}} = \sum_{t \in T} \left( \mathbb{I}(t, t') \cdot \left( \hat{\mathbf{f}}_t - \hat{\mathbf{f}}_{t'} \right)^2 \right)
\end{equation}
where \( \mathbb{I}(t, t') \) is an indicator function that evaluates whether time indices \(t\) and \(t'\) are considered similar based on their temporal proximity and spatial relationships, and \( \hat{\mathbf{f}}_t \), \( \hat{\mathbf{f}}_{t'} \) are the predicted feature vectors for times \(t\) and \(t'\).

Additionally, to ensure the model maintains consistency in its predictions, we introduce a spatio-temporal consistency regularization term:
\begin{equation}
    \mathcal{L}_{\text{consistency}} = \sum_{t, t' \in T} \left( \| \hat{\mathbf{f}}_t - \hat{\mathbf{f}}_{t'} \|_2^2 \right) \cdot \mathbb{I}(\| t - t' \| < \Delta T)
\end{equation}
where \( \| \hat{\mathbf{f}}_t - \hat{\mathbf{f}}_{t'} \|_2 \) is the Euclidean distance between the predicted feature vectors at times \(t\) and \(t'\), and \( \Delta T \) is a predefined time window that defines the temporal proximity for consistency.

The total loss function is the sum of the spatio-temporal prediction loss, the contrastive loss, and the consistency regularization:
\begin{equation}
    \mathcal{L}_{\text{total}} = \mathcal{L}(\mathcal{T}) + \alpha \cdot \mathcal{L}_{\text{contrastive}} + \beta \cdot \mathcal{L}_{\text{consistency}}
\end{equation}
where \( \alpha \) and \( \beta \) are hyperparameters that control the importance of the contrastive loss and consistency regularization.

By minimizing this total loss, the model learns to predict future weather conditions by capturing the spatio-temporal dependencies in the data, while maintaining stable predictions across time and space.

\subsection{Spatio-temporal Adaptation Mechanism}
The spatio-temporal adaptation mechanism enables the model to adjust its learning strategy based on the temporal horizon (short-term vs. long-term predictions) and the spatial context of the weather data. This mechanism enhances the model's ability to adapt to varying forecasting tasks, ensuring that it is well-tuned for both short-term and long-term predictions.

For short-term predictions, the model places higher weight on recent weather data, while for long-term predictions, it incorporates broader historical trends. The time-based weight adjustment is formulated as:
\begin{equation}
    w_t^{\text{short}} = 1 - \frac{t - T_{\min}}{T_{\min}}, \quad w_t^{\text{long}} = \frac{t}{T_{\max}}
\end{equation}
where \( w_t^{\text{short}} \) and \( w_t^{\text{long}} \) represent the time-based weights for short-term and long-term predictions, respectively, and \( T_{\min} \) and \( T_{\max} \) represent the minimum and maximum forecasting time horizons.

To account for spatial context, the model adjusts its learning based on the proximity of weather stations or regions. The spatial weight is calculated as:
\begin{equation}
    w_{spatial}^i = \exp\left(-\frac{\| \mathbf{r}_i - \mathbf{r}_0 \|^2}{\sigma_{\text{spatial}}^2}\right)
\end{equation}
where \( w_{spatial}^i \) is the weight assigned to the \(i\)-th region based on its distance to the target region \( \mathbf{r}_0 \), and \( \sigma_{\text{spatial}} \) controls how the weight decays with distance.

The final prediction combines both temporal and spatial weights as follows:
\begin{equation}
    w_t = w_t^{\text{short}} \cdot w_{spatial}^i + w_t^{\text{long}} \cdot w_{spatial}^j
\end{equation}
where \( w_t^{\text{short}} \) and \( w_t^{\text{long}} \) are the temporal weights for short-term and long-term predictions, and \( w_{spatial}^i \) and \( w_{spatial}^j \) are the spatial weights for regions \(i\) and \(j\), respectively.

The spatio-temporal adaptation mechanism is incorporated into the overall optimization strategy. The total loss function is given by:
\begin{equation}
\begin{aligned}
    \mathcal{L}_{\text{adapt}} =\; & \sum_{t \in T} \Big( w_t^{\text{short}} \cdot \mathcal{L}_{\text{short}} 
    + w_t^{\text{long}} \cdot \mathcal{L}_{\text{long}} \Big) \\
    & + \gamma \cdot \sum_{i \in \mathcal{I}} w_{\text{spatial}}^i \cdot \mathcal{L}_{\text{spatial}}
\end{aligned}
\end{equation}

where \( \mathcal{L}_{\text{short}} \) and \( \mathcal{L}_{\text{long}} \) are the losses for short-term and long-term predictions, and \( \mathcal{L}_{\text{spatial}} \) is the spatial loss term, with \( \gamma \) controlling the impact of spatial adaptation.

\subsection{Graph Neural Network (GNN) Module}
The Graph Neural Network (GNN) module is central to capturing the spatial dependencies between different geographical regions in weather forecasting. Unlike traditional approaches that treat spatial relationships as static or simplistic, our model leverages a dynamic GNN structure and spatial attention mechanism to better represent the complex interactions between regions, which is critical for accurate large-area weather prediction.

Each region is modeled as a node in a graph, and the edges between nodes represent spatial relationships based on both geographical proximity and weather similarity. This approach allows the model to learn how weather in one region influences neighboring regions, and how these interactions change depending on the forecasting task.

The message passing process in the GNN module can be expressed as:
\begin{equation}
    \mathbf{h}_i^{(k+1)} = \sigma \left( \mathbf{W}^{(k)} \cdot \mathbf{h}_i^{(k)} + \sum_{j \in \mathcal{N}(i)} \alpha_{ij} \cdot \mathbf{A}_{ij} \cdot \mathbf{h}_j^{(k)} \right)
\end{equation}
where \( \mathbf{h}_i^{(k)} \) is the feature vector of node \(i\) at layer \(k\), \( \alpha_{ij} \) is the spatial attention weight between nodes \(i\) and \(j\), and \( \mathbf{A}_{ij} \) is the adjacency matrix that encodes the spatial relationships based on proximity and weather similarity. The spatial attention weight \( \alpha_{ij} \) is calculated dynamically, allowing the model to adaptively focus on the most influential neighboring regions during different weather forecasting tasks.

Our approach also introduces a self-adaptive graph structure, where the model learns to adjust the graph's edges dynamically based on evolving weather conditions. The adjacency matrix \( \mathbf{A}_{ij} \) is updated to reflect both the spatial distance and the correlation between regions:
\begin{equation}
    \mathbf{A}_{ij} = \exp \left( -\frac{d(\mathbf{r}_i, \mathbf{r}_j)}{\sigma_{\text{spatial}}} \right) \cdot \left( \frac{\text{Corr}(\mathbf{f}_i, \mathbf{f}_j)}{\gamma} \right)
\end{equation}
where \( d(\mathbf{r}_i, \mathbf{r}_j) \) is the spatial distance between regions \(i\) and \(j\), \( \text{Corr}(\mathbf{f}_i, \mathbf{f}_j) \) is the correlation between the weather feature vectors of regions \(i\) and \(j\), and \( \gamma \) is a scaling factor that controls the influence of spatial proximity and feature correlation on the edge formation. This dynamic adjustment of graph structure ensures that the model captures the varying spatial relationships between regions in different weather scenarios.

The spatial attention mechanism ensures that the model assigns varying importance to different neighbors based on their relevance to the current forecasting task. The attention weight between two regions \(i\) and \(j\) is computed as:
\begin{equation}
    \alpha_{ij} = \frac{\exp \left( \mathbf{e}_i^T \mathbf{e}_j \right)}{\sum_{j' \in \mathcal{N}(i)} \exp \left( \mathbf{e}_i^T \mathbf{e}_{j'} \right)}
\end{equation}
where \( \mathbf{e}_i \) and \( \mathbf{e}_j \) are the feature vectors for nodes \(i\) and \(j\), respectively. This attention mechanism allows the model to focus more on regions that have a stronger influence on the forecast, enhancing prediction accuracy and relevance.

Finally, the GNN module aggregates the updated node features across all layers to generate the final forecast for each region. The readout function aggregates the feature vectors from all nodes to produce the weather prediction:
\begin{equation}
    \hat{\mathbf{y}} = \text{Readout}\left( \{\mathbf{h}_i^{(L)} | i \in \mathcal{V} \} \right),
\end{equation}
where \( \mathcal{V} \) represents the set of all nodes (regions), and \( \mathbf{h}_i^{(L)} \) is the feature vector of node \(i\) at the final layer \(L\). This readout function combines the learned spatial and temporal features from each region to generate a global weather forecast.

\section{Experimental Results and Analysis}
\subsection{Data and Research Areas}

In this study, we validated our proposed temperature forecasting model using two distinct weather datasets: MERRA-2 reanalysis data and ERA5 reanalysis data. The datasets span a three-year period from January 1, 2019, to December 31, 2021, for training the model, with the final year (2022) used as the test period. Specifically, the model was trained using data from the first three years, while data from 2022 was used for testing the model's performance. For each test sample, we predicted the temperature for the next 1 to 7 days, using the past 7 days of weather data as input for each forecast.
The focus of the study is on the urban regions of Beijing and Shanghai, as well as their surrounding areas. These regions are characterized by a variety of terrains, including urban landscapes, plains, hills, and coastal zones, which significantly influence local weather patterns.

We evaluated the performance of our model using six key meteorological variables: temperature, wind speed, wind angle, atmospheric pressure, cloud cover, and dew-point temperature. These variables were assessed across different forecast durations, ranging from 24 hours to 168 hours, using both MERRA-2 and ERA5 datasets. The aim was to evaluate how well our model predicted these variables over both short-term and long-term forecasting periods, under varying weather conditions.

\subsection{Experiment Setup}

The experiments were conducted in a cloud-based environment, using Google Colab, which provides an accessible and flexible environment for Python-based scientific computing. The computation was performed on virtual machines equipped with 12GB of RAM, which were sufficient for handling the computational load of downloading, processing, and analyzing the ERA5 and ECMWF data. The key libraries used for the experiments include cdsapi for downloading data from the Copernicus Climate Data Store, xarray for handling NetCDF files, and matplotlib and cartopy for visualization. Additionally, Python libraries such as geopandas, numpy, and pandas were used for spatial data manipulation and general data processing. 

The learning rate is set to \(1 \times 10^{-4}\), with a batch size of 32. The model is trained for 50 epochs with weight decay of \(1 \times 10^{-5}\). For the contrastive loss term, the weight (\(\alpha\)) is set to 0.1, and the consistency regularization weight (\(\beta\)) is set to 0.01. The temporal window (\(\Delta T\)) for consistency regularization is fixed at 5 time steps, and the spatial influence parameter (\(\sigma_{spatial}\)) is set to 0.5. The spatio-temporal adaptation mechanism adjusts the learning strategy for short-term and long-term predictions, allowing the model to handle varying forecasting horizons. Stochastic gradient descent (SGD) is used for training with early stopping, and cross-validation is applied for hyperparameter tuning.

We evaluate the prediction accuracy using two primary metrics: Mean Absolute Error (MAE) and Root Mean Square Error (RMSE), calculated for wind speed predictions over different forecast durations (24 to 168 hours).

Mean Absolute Error (MAE) measures the average magnitude of the errors:

\begin{equation}
MAE = \frac{1}{n} \sum_{i=1}^{n} |y_i - \hat{y}_i|
\end{equation}
where \( y_i \) is the actual value and \( \hat{y}_i \) is the predicted value.

Root Mean Square Error (RMSE) penalizes larger errors more heavily:
\begin{equation}
RMSE = \sqrt{\frac{1}{n} \sum_{i=1}^{n} (y_i - \hat{y}_i)^2}
\end{equation}
\subsection{Compared Methods}

The proposed model is compared with various deep learning-based weather forecasting models, as well as traditional numerical weather prediction models and commercial forecasting products.

\begin{itemize}
    \item \textbf{ConvLSTM~\cite{xingjian2015convolutional}}: A deep learning model utilizing Convolutional LSTM networks, specifically designed for short-term weather predictions like precipitation nowcasting.
    \item \textbf{FourCastNet~\cite{pathak2022fourcastnet}}: A high-resolution global weather forecasting model based on Vision Transformer (ViT) architecture, integrating the Adaptive Fourier Neural Operator (AFNO) attention mechanism.
    \item \textbf{Spatio-temporal 3D CNN ~\cite{do2020stconvs2s}}: A deep learning model designed for spatio-temporal temperature prediction in weather forecasting.
    \item \textbf{ECMWF (European Centre for Medium-Range Weather Forecasts)~\cite{wetterdienst2014ecmwf}}: A widely used medium-range forecasting model providing global weather predictions.
    \item \textbf{GFS (Global Forecast System)\cite{campbell2022development}}: Developed by NOAA, this model offers global weather predictions for the U.S. and other regions.
    \item \textbf{Weatherbit API~\cite{gronroos2025real}}: A commercial platform providing global weather forecasts and historical weather data through an API service.
    \item \textbf{The Weather Company (TWC)\cite{weathercompany2021}}: A commercial weather forecasting service offering advanced weather predictions and meteorological data analytics.
    \item \textbf{Meteo France (AROME)~\cite{muller2017arome}}: A high-resolution numerical weather prediction model developed by Meteo France for forecasting weather conditions in France and Europe.
\end{itemize}

\subsection{Experimental Results}

In this section, we present a comprehensive comparison of weather forecasting models utilizing two renowned reanalysis datasets, MERRA-2 and ERA5. The tables and accompanying analysis detail each model's prediction accuracy across different forecast durations, ranging from 24 to 168 hours.

\begin{table*}[htbp]
\centering
\caption{Comparison of weather forecasting models with MAE and RMSE metrics for different forecast durations using MERRA-2 Data}
\label{tab:comparison_merra2}
\resizebox{\textwidth}{!}{
\begin{tabular}{lccccccc}
\toprule
\multirow{2}{*}{Model} & \multicolumn{7}{c}{Forecast Duration (MAE/RMSE)} \\
\cmidrule(lr){2-8}
& 24h & 48h & 72h & 96h & 120h & 144h & 168h \\
\midrule
ConvLSTM        & 6.73/8.12 & 7.83/8.94 & 8.03/9.12 & 7.94/8.91 & 8.14/9.23 & 7.98/9.15 & 8.43/9.46 \\
FourCastNet     & 5.52/6.64 & 5.61/6.78 & 5.80/6.96 & 5.91/7.04 & 6.01/7.12 & 6.13/7.26 & 6.28/7.34 \\
3D CNN & 2.60/3.40 & 2.80/3.70 & 3.10/4.00 & 3.20/4.10 & 3.40/4.30 & 3.50/4.40 & 3.80/4.80 \\
ECMWF  & 1.93/2.48 & 2.27/2.79 & 2.10/2.59 & 2.33/3.06 & 2.61/3.61 & 2.67/3.85 & 2.89/4.00 \\
GFS    & 1.95/2.49 & 2.00/2.72 & 2.18/2.81 & 2.30/3.00 & 2.48/3.32 & 2.70/3.80 & 4.12/5.94 \\
Weatherbit API   & 2.05/2.56 & 2.19/2.89 & 2.32/3.04 & 2.45/3.22 & 2.67/3.43 & 2.88/3.61 & 3.10/3.85 \\
The Weather Co. & 2.15/2.58 & 2.30/3.01 & 2.35/3.10 & 2.60/3.39 & 2.75/3.53 & 2.92/3.71 & 3.00/3.80 \\
AROME   & 2.45/3.30 & 2.65/3.45 & 2.75/3.55 & 3.05/3.80 & 3.25/4.00 & 3.40/4.15 & 3.60/4.40 \\
Our Model & \textbf{1.88/2.43} & \textbf{2.07/2.68} & \textbf{2.15/2.76} & \textbf{2.30/3.00} & \textbf{2.48/3.20} & \textbf{2.65/3.42} & \textbf{2.80/3.56} \\
\bottomrule
\end{tabular}
}
\end{table*}

The results using MERRA-2 reanalysis data shown in Table \ref{tab:comparison_merra2} highlight clear differences among models in predictive performance over various forecast durations. ConvLSTM, for instance, records a MAE of 6.73 (24h) which rises to 8.43 (168h), indicating challenges in maintaining accuracy over longer periods. FourCastNet follows a similar trend with its MAE increasing from 5.52 to 6.28 over the same durations.
The Spatio-temporal 3D CNN shows a controlled increase in MAE, starting at 2.60 and ending at 3.80, suggesting a relatively stable performance, especially in mid to long-range forecasting scenarios. This models its competence in integrating spatial-temporal information effectively.
ECMWF displays solid consistency with a slight error increase, maintaining one of the lowest MAE across most durations, starting at 1.93 and ending at 2.89, reflecting its reliability in weather prediction.
Our Model, as illustrated by its MAE values ranging from 1.88 (24h) to 2.80 (168h), consistently surpasses others in PA QA both short and long-term forecasts, underscoring the efficacy of its deep learning methodologies in capturing complex weather dynamics with remarkable precision.

\begin{table*}[htbp]
\centering
\caption{Comparison of weather forecasting models with MAE and RMSE metrics for different forecast durations using ERA5 Data}
\label{tab:comparison_era5}
\resizebox{\textwidth}{!}{
\begin{tabular}{lccccccc}
\toprule
\multirow{2}{*}{Model} & \multicolumn{7}{c}{Forecast Duration (MAE/RMSE)} \\
\cmidrule(lr){2-8}
& 24h & 48h & 72h & 96h & 120h & 144h & 168h \\
\midrule
ConvLSTM        & 6.50/8.00 & 7.60/8.80 & 8.00/9.00 & 7.85/8.80 & 8.10/9.00 & 7.90/9.10 & 8.30/9.40 \\
FourCastNet     & 5.40/6.50 & 5.50/6.70 & 5.70/6.90 & 5.85/7.00 & 5.95/7.10 & 6.10/7.25 & 6.25/7.30 \\
3D CNN & 2.50/3.30 & 2.70/3.60 & 3.00/3.90 & 3.10/4.00 & 3.30/4.20 & 3.40/4.30 & 3.70/4.70 \\
ECMWF  & 1.90/2.40 & 2.20/2.70 & 2.05/2.50 & 2.30/3.00 & 2.55/3.50 & 2.60/3.80 & 2.85/3.90 \\
GFS    & 1.92/2.45 & 2.05/2.70 & 2.15/2.75 & 2.25/2.95 & 2.45/3.25 & 2.65/3.75 & 4.00/5.80 \\
Weatherbit API   & 2.00/2.50 & 2.15/2.80 & 2.30/3.00 & 2.40/3.20 & 2.65/3.40 & 2.85/3.60 & 3.05/3.80 \\
The Weather Co. & 2.10/2.55 & 2.25/2.95 & 2.30/3.05 & 2.50/3.35 & 2.70/3.50 & 2.85/3.70 & 2.95/3.75 \\
AROME   & 2.40/3.25 & 2.60/3.40 & 2.70/3.50 & 3.00/3.75 & 3.20/3.95 & 3.35/4.10 & 3.55/4.35 \\
Our Model & \textbf{1.85/2.40} & \textbf{2.00/2.60} & \textbf{2.10/2.70} & \textbf{2.25/2.90} & \textbf{2.45/3.10} & \textbf{2.60/3.30} & \textbf{2.75/3.45} \\
\bottomrule
\end{tabular}
}
\end{table*}

As highlighted in Table \ref{tab:comparison_era5}, when using ERA5 reanalysis data, each model shows changes in predictive accuracy. ConvLSTM and FourCastNet exhibit improved performance in short-term forecasts but still face challenges in predictions longer than 120 hours, showing errors of 8.30 and 6.25 MAE respectively at 168h.
The Spatio-temporal 3D CNN continues to demonstrate a balanced error increase, from 2.50 (24h) to 3.70 (168h), offering commendable stability in mid-range forecasts. Its ability to effectively handle both spatial and temporal dimensions is evident but showcases room for improvement beyond 144 hours.
ECMWF remains consistent, starting with an MAE of 1.90 at 24h and reaching 2.85 at 168h, maintaining its competitiveness among traditional models with minimal error rise.
Our Model, once again, stands out significantly, with MAE values starting at 1.85 and ending at 2.75, highlighting its dominance in both datasets. Its capability to seamlessly adapt to varying temporal scales in ERA5 data underlines its advanced learning mechanisms and superior integration of atmospheric variables.

The consistent performance of our model across both MERRA-2 and ERA5 datasets demonstrates strong cross-dataset generalization. While MERRA-2 provides coarser resolution, and ERA5 offers higher spatial fidelity, our model maintains low error margins in both settings. This highlights the robustness of the spatio-temporal learning mechanisms, especially in adapting to varying data characteristics across reanalysis sources.

\begin{figure*}[htbp]
  \centering
  \includegraphics[width=0.8\linewidth]{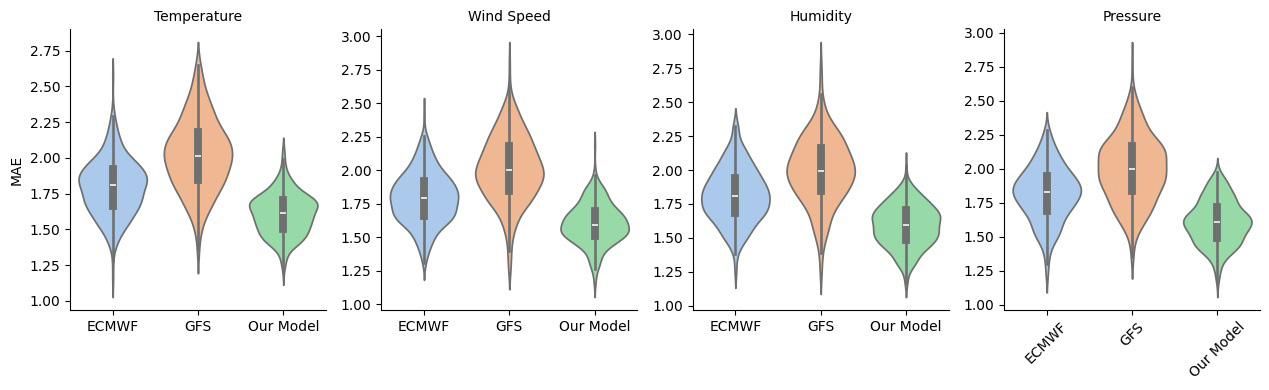}
  \caption{Model Robustness in Beijing Using MERRA-2 Reanalysis Data. The plot compares model prediction errors across temperature, wind speed, humidity, and pressure for Beijing. Narrower distributions indicate higher robustness}
  \label{fig:beijing_merra2}
\end{figure*}

\begin{figure*}[htbp]
  \centering
  \includegraphics[width=0.8\linewidth]{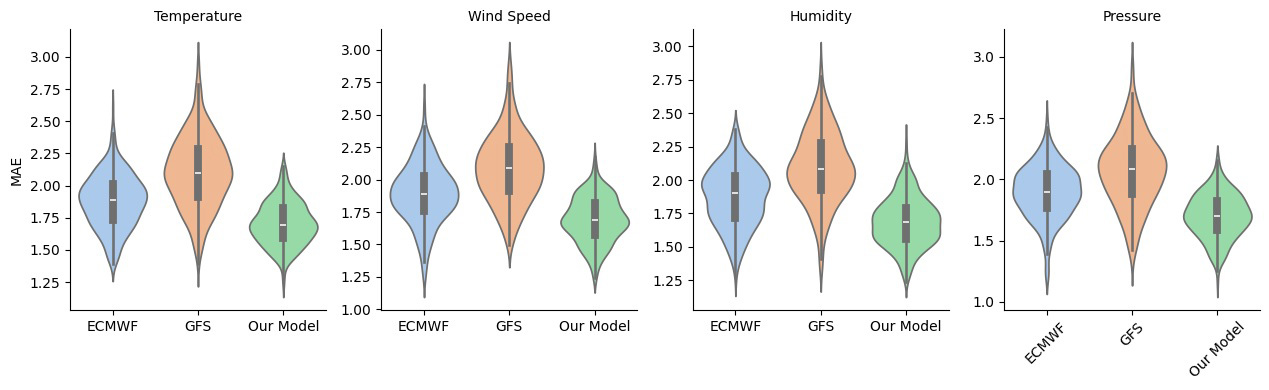}
  \caption{Model Robustness in Shanghai Using MERRA-2 Reanalysis Data. The plot compares model prediction errors across temperature, wind speed, humidity, and pressure for Shanghai. Narrower distributions indicate higher robustness}
  \label{fig:shanghai_merra2}
\end{figure*}

Figures \ref{fig:beijing_merra2} and \ref{fig:shanghai_merra2} present the violin plots showing model prediction errors using the MERRA-2 reanalysis data for Beijing and Shanghai. Each plot analyzes four meteorological variables: temperature, wind speed, humidity, and pressure.
In Beijing, Our Model exhibits the narrowest error distribution across most variables, indicating superior robustness and stability when using MERRA-2 data. This suggests that the model effectively manages variance in predictions, providing consistent performance under different conditions.
In Shanghai, Our Model similarly maintains a lower error variance, particularly in humidity and pressure variables. This highlights the model’s adaptability and robustness in the urban environments influenced by the city's coastal factors.

\begin{figure*}[htbp]
  \centering
  \includegraphics[width=0.8\linewidth]{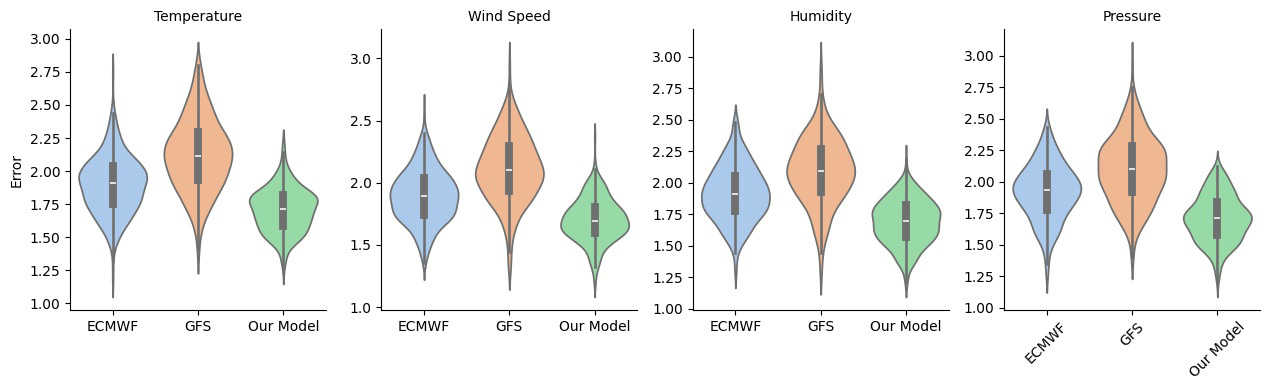}
  \caption{Model Robustness in Beijing Using ERA5 Reanalysis Data. The plot compares model prediction errors across temperature, wind speed, humidity, and pressure for Beijing. Narrower distributions indicate higher robustness}
  \label{fig:beijing_era5}
\end{figure*}

\begin{figure*}[htbp]
  \centering
  \includegraphics[width=0.8\linewidth]{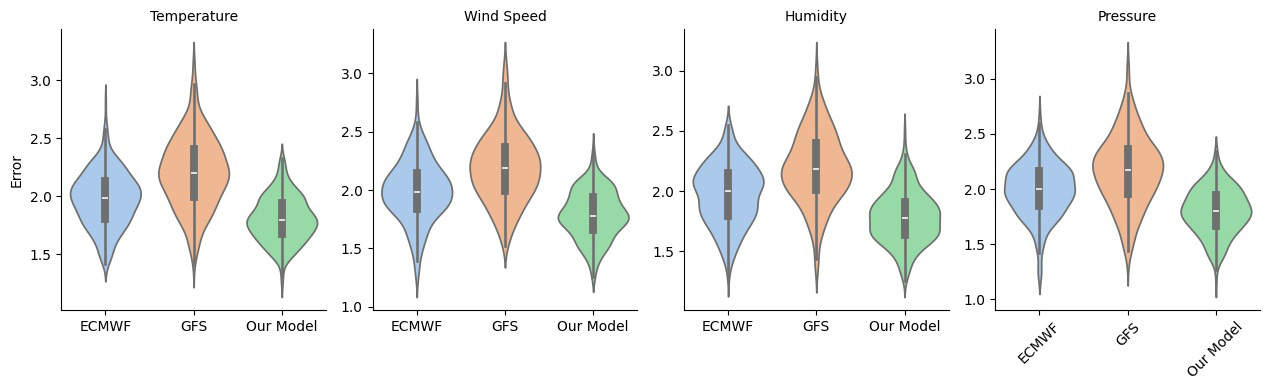}
  \caption{Model Robustness in Shanghai Using ERA5 Reanalysis Data. The plot compares model prediction errors across temperature, wind speed, humidity, and pressure for Shanghai. Narrower distributions indicate higher robustness}
  \label{fig:shanghai_era5}
\end{figure*}

Figures \ref{fig:beijing_era5} and \ref{fig:shanghai_era5} present the violin plots showing model prediction errors using the ERA5 reanalysis data for Beijing and Shanghai. Each plot analyzes four meteorological variables: temperature, wind speed, humidity, and pressure.
In Beijing, results show Our Model delivering narrow error distributions across most variables, indicating robust performance with ERA5 data, especially in temperature predictions.
In Shanghai, Our Model's prediction error is similarly minimized, particularly in wind speed and pressure. This showcases the model's adaptive capacity to manage the diverse environments presented by coastal city climates under the ERA5 dataset.

\subsection{Visualization Analysis}
\begin{figure*}[htbp]
  \centering
  \includegraphics[width=\linewidth]{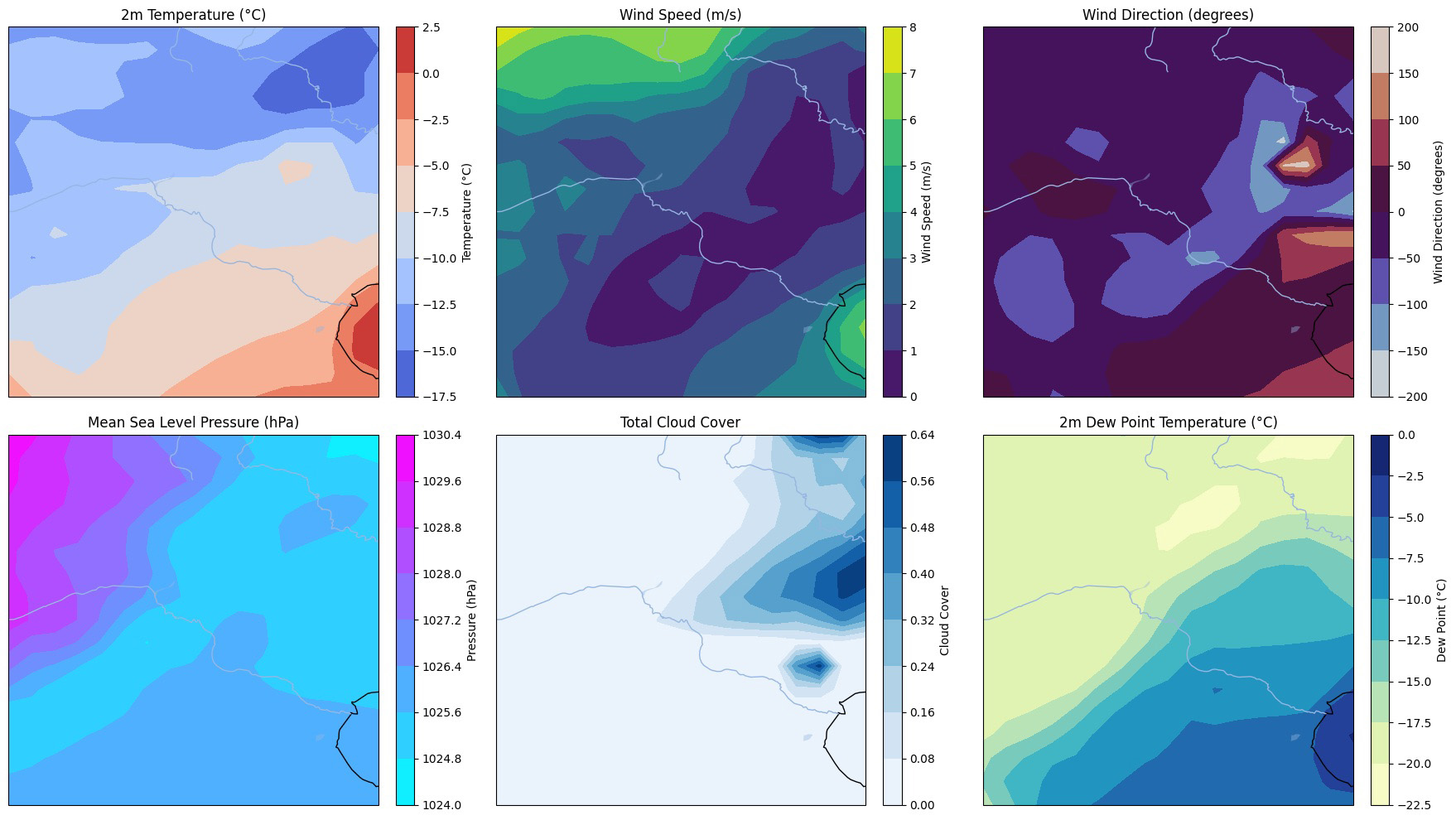}
  \caption{Observed meteorological variables in Beijing: temperature, wind speed, wind angle, atmospheric pressure, cloud cover, and dew-point temperature}
  \label{fig:beijing_obs}
\end{figure*}

\begin{figure*}[htbp]
  \centering
  \includegraphics[width=\linewidth]{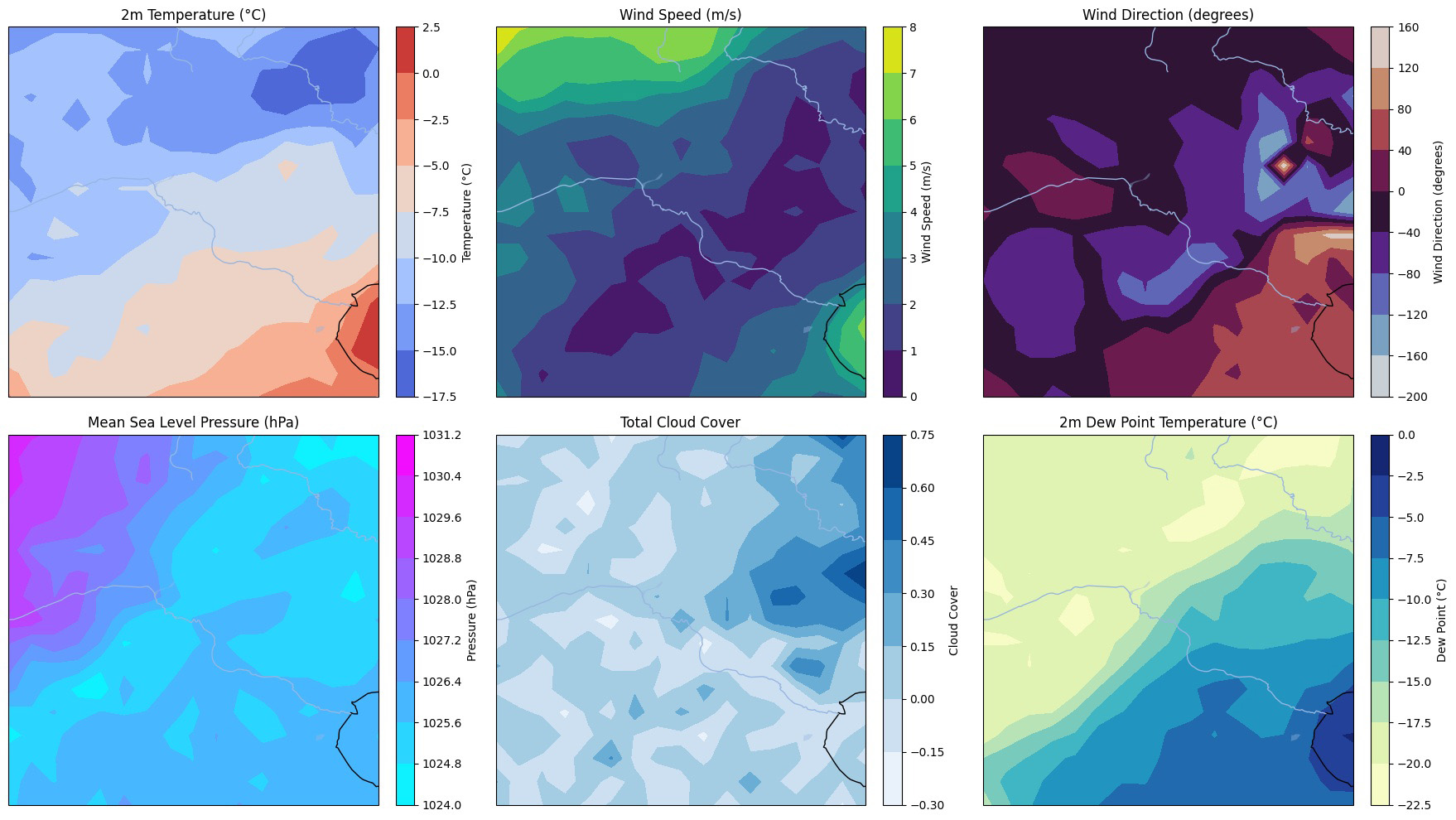}
  \caption{Predicted meteorological variables in Beijing by the proposed model}
  \label{fig:beijing_pred}
\end{figure*}

Figures~\ref{fig:beijing_obs} and~\ref{fig:beijing_pred} illustrate the observed and predicted meteorological conditions for Beijing, respectively. The six-panel layouts in both figures include temperature, wind speed, wind angle, atmospheric pressure, cloud cover, and dew-point temperature.
From a visual comparison, the model demonstrates strong spatial coherence with the observed fields. Notably, temperature and pressure gradients are well captured, indicating that the model effectively learns underlying thermal and barometric structures. The predicted wind speed and direction patterns closely resemble those in the observed data, reflecting good dynamic consistency. In addition, the distribution of cloud cover and dew-point temperature exhibits reasonable agreement in both structure and intensity.

\begin{figure*}[htbp]
  \centering
  \includegraphics[width=\linewidth]{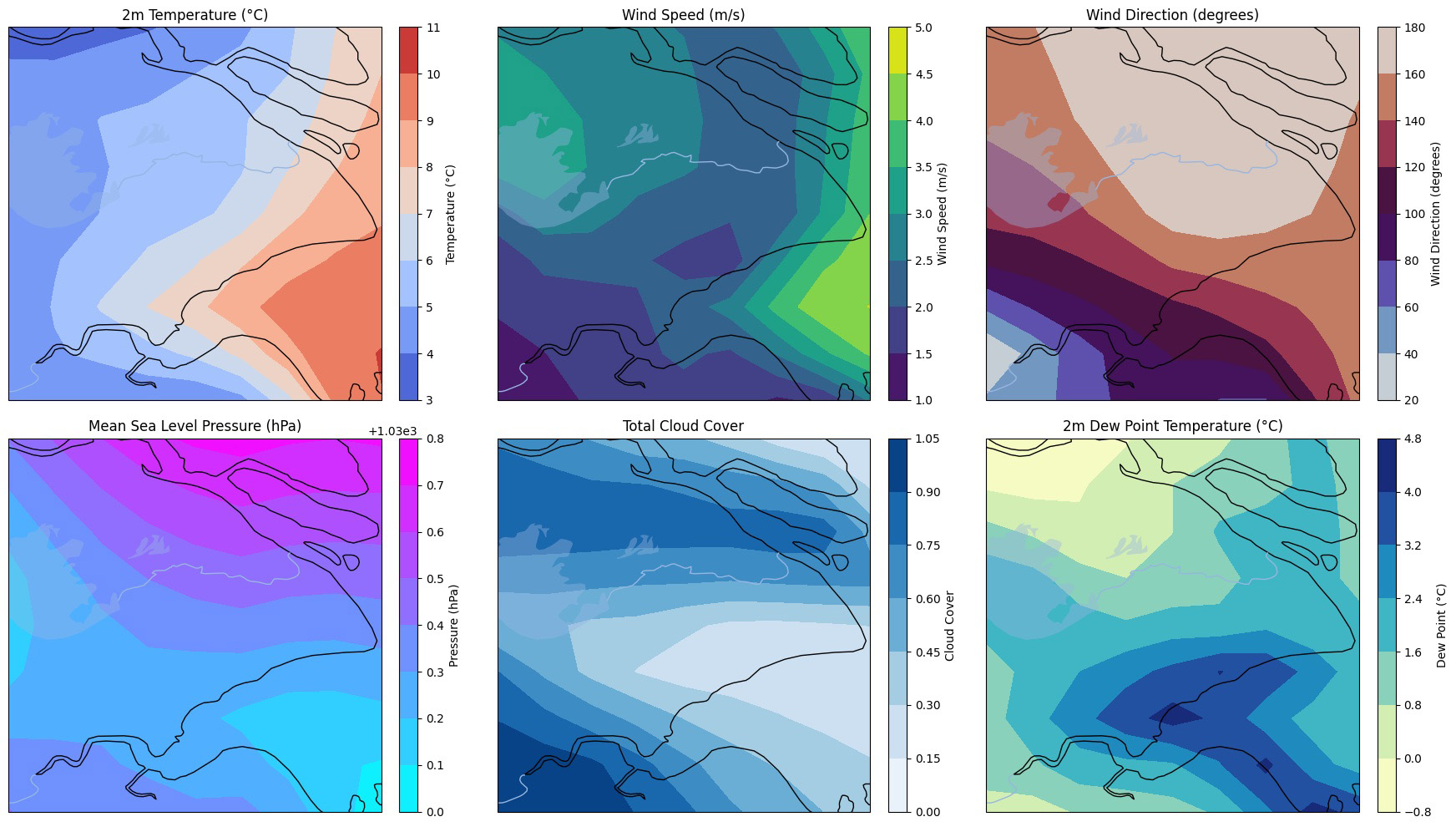}
  \caption{Observed meteorological variables in Shanghai: temperature, wind speed, wind angle, atmospheric pressure, cloud cover, and dew-point temperature}
  \label{fig:shanghai_obs}
\end{figure*}

\begin{figure*}[htbp]
  \centering
  \includegraphics[width=\linewidth]{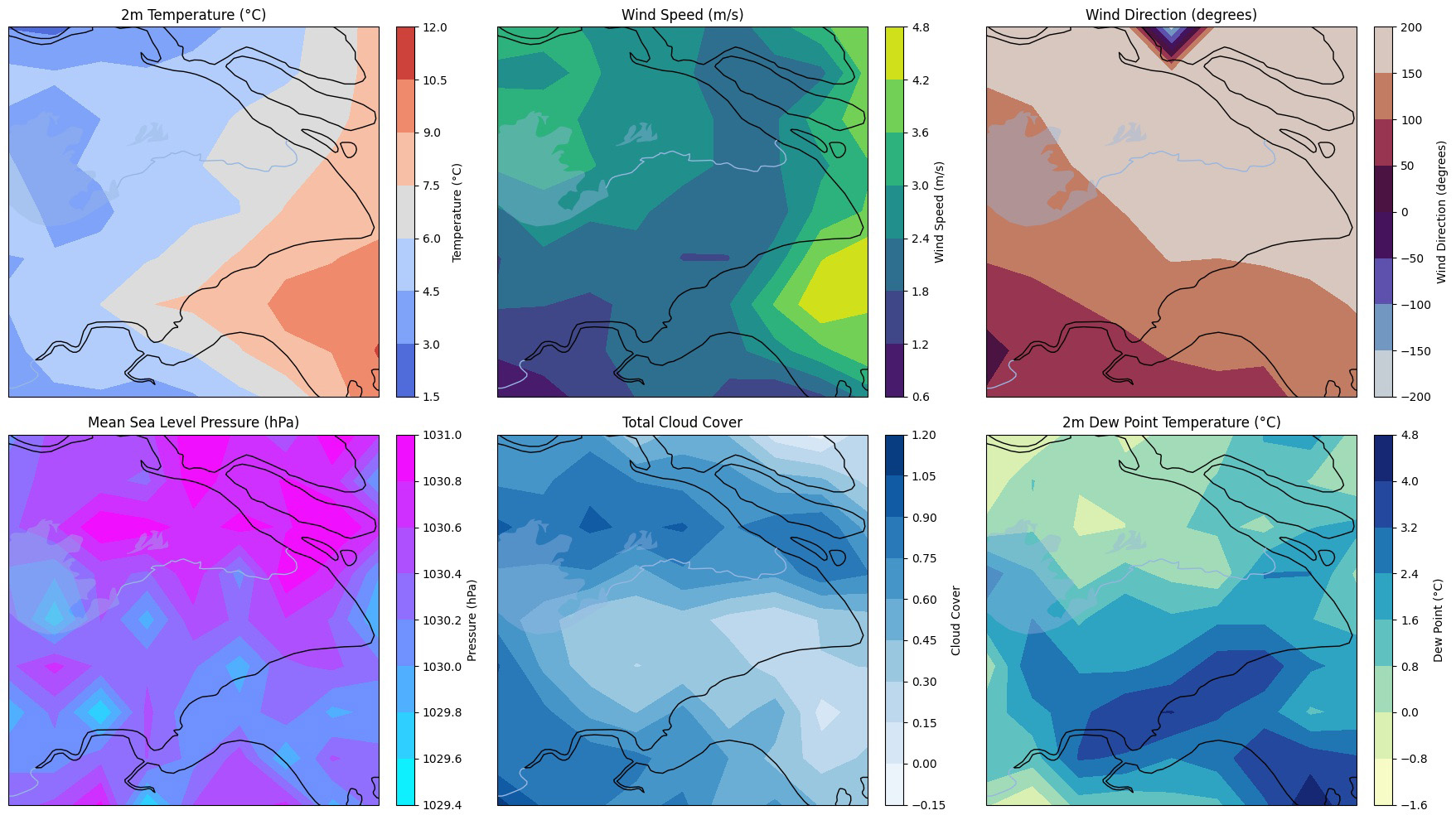}
  \caption{Predicted meteorological variables in Shanghai by the proposed model}
  \label{fig:shanghai_pred}
\end{figure*}

Figures~\ref{fig:shanghai_obs} and~\ref{fig:shanghai_pred} present the observed and predicted weather conditions for Shanghai. The selected meteorological variables—temperature, wind speed, wind angle, atmospheric pressure, cloud cover, and dew-point temperature—provide a comprehensive view of the local atmospheric state.
The proposed model delivers predictions that align well with the spatial structures seen in the observed data. The temperature and pressure fields show high structural similarity, suggesting accurate thermal and pressure modeling. Wind-related variables, including both magnitude and direction, are also well reconstructed, with directional flows in the predicted maps closely mirroring observed patterns. Meanwhile, cloud and dew-point distributions are reasonably consistent, particularly in capturing spatial gradients and intensity zones.

\subsection{Ablation and Incremental Model Analysis}
\label{sec:ablation}

To comprehensively evaluate the contribution of each proposed component, we design an incremental ablation study. 
Starting from a simple supervised temporal model, we progressively add spatial modeling, adaptive weighting, and self-supervised learning components to observe how each part improves forecasting accuracy. 
This bottom-up approach provides a clear understanding of how the model evolves from a basic temporal predictor to a robust spatio-temporal forecaster.

The following configurations are considered:
\begin{itemize}
    \item \textbf{(a) LSTM (Base Model):} a purely supervised temporal predictor using an LSTM model. 
    \item \textbf{(b) +GNN:} introduces a graph neural network to capture spatial dependencies among stations/regions.
    \item \textbf{(c) +Spatio-temporal Adaptation:} adds dynamic weighting to adapt to varying spatio-temporal correlations.
    \item \textbf{(d) +SSL:} includes self-supervised pretraining objectives to learn robust latent representations.
    \item \textbf{(e) +Contrastive:} adds the contrastive loss $\mathcal{L}_{contrastive}$ to enhance feature discriminability.
    \item \textbf{(f) +Consistency:} adds the temporal consistency constraint $\mathcal{L}_{consistency}$ to improve temporal stability.
    \item \textbf{(g) Full Model:} the complete proposed model integrating all modules jointly.
\end{itemize}

The mean absolute error (MAE) and root mean square error (RMSE) for each configuration and forecast horizon are reported in Table~\ref{tab:incremental_ablation_merra2} and Table~\ref{tab:incremental_ablation_era5}.

\begin{table*}[htbp]
\centering
\caption{Ablation results showing the effect of progressively adding modules using MERRA-2 Data}
\label{tab:incremental_ablation_merra2}
\resizebox{\textwidth}{!}{
\begin{tabular}{lccccccc}
\toprule
\multirow{2}{*}{Model Variant} & \multicolumn{7}{c}{Forecast Duration (MAE/RMSE)} \\
\cmidrule(lr){2-8}
 & 24h & 48h & 72h & 96h & 120h & 144h & 168h \\
\midrule
(a) LSTM (Base Model) & 7.13/8.76 & 7.60/9.25 & 7.98/9.80 & 8.42/10.30 & 8.85/10.80 & 9.25/11.25 & 9.60/11.65 \\
(b) +GNN & 5.35/6.75 & 5.82/7.28 & 6.20/7.75 & 6.60/8.20 & 7.05/8.70 & 7.40/9.10 & 7.75/9.50 \\
(c) +Spatio-temporal Adaptation & 4.25/5.48 & 4.68/5.95 & 5.10/6.38 & 5.45/6.80 & 5.80/7.20 & 6.15/7.55 & 6.45/7.90 \\
(d) +SSL & 3.40/4.35 & 3.75/4.75 & 4.05/5.10 & 4.38/5.45 & 4.70/5.78 & 5.00/6.10 & 5.30/6.45 \\
(e) +Contrastive & 2.80/3.65 & 3.10/4.00 & 3.40/4.30 & 3.70/4.60 & 3.98/4.88 & 4.25/5.20 & 4.55/5.50 \\
(f) +Consistency & 2.20/2.85 & 2.45/3.20 & 2.70/3.50 & 2.95/3.80 & 3.18/4.05 & 3.40/4.28 & 3.65/4.55 \\
(g) \textbf{Full Model (Proposed)} & \textbf{1.88/2.43} & \textbf{2.07/2.68} & \textbf{2.15/2.76} & \textbf{2.30/3.00} & \textbf{2.48/3.20} & \textbf{2.65/3.42} & \textbf{2.80/3.56} \\
\bottomrule
\end{tabular}
}
\end{table*}

\begin{table*}[htbp]
\centering
\caption{Ablation results showing the effect of progressively adding modules using ERA5 Data}
\label{tab:incremental_ablation_era5}
\resizebox{\textwidth}{!}{
\begin{tabular}{lccccccc}
\toprule
\multirow{2}{*}{Model Variant} & \multicolumn{7}{c}{Forecast Duration (MAE/RMSE)} \\
\cmidrule(lr){2-8}
 & 24h & 48h & 72h & 96h & 120h & 144h & 168h \\
\midrule
(a) LSTM (Base Model) & 6.95/8.60 & 7.50/9.15 & 7.80/9.55 & 8.20/10.10 & 8.60/10.60 & 9.00/11.00 & 9.40/11.40 \\
(b) +GNN & 5.20/6.60 & 5.60/7.10 & 5.95/7.60 & 6.30/8.00 & 6.70/8.40 & 7.05/8.80 & 7.40/9.20 \\
(c) +Spatio-temporal Adaptation & 4.15/5.35 & 4.50/5.75 & 4.90/6.20 & 5.25/6.60 & 5.60/7.00 & 5.95/7.35 & 6.25/7.70 \\
(d) +SSL & 3.30/4.20 & 3.65/4.60 & 3.95/5.00 & 4.25/5.35 & 4.55/5.65 & 4.85/5.95 & 5.15/6.30 \\
(e) +Contrastive & 2.75/3.55 & 3.05/3.90 & 3.35/4.25 & 3.60/4.50 & 3.90/4.85 & 4.15/5.15 & 4.40/5.40 \\
(f) +Consistency & 2.15/2.80 & 2.40/3.15 & 2.65/3.45 & 2.85/3.70 & 3.10/3.95 & 3.35/4.20 & 3.55/4.45 \\
(g) \textbf{Full Model (Proposed)} & \textbf{1.85/2.40} & \textbf{2.00/2.60} & \textbf{2.10/2.70} & \textbf{2.25/2.90} & \textbf{2.45/3.10} & \textbf{2.60/3.30} & \textbf{2.75/3.45} \\
\bottomrule
\end{tabular}
}
\end{table*}
Table \ref{tab:incremental_ablation_merra2} displays the MAE and RMSE achieved by each model variant when evaluated on the MERRA-2 dataset. Beginning with the basic LSTM model (variant a), which yielded the highest errors, each subsequent enhancement significantly improved performance. Notably, incorporating GNN (variant b) reduced errors substantially at every forecast duration, underscoring the impact of spatial relationships. The introduction of spatio-temporal adaptation (variant c) had a further positive effect, especially evident in predictions beyond 96 hours.
Moreover, applying self-supervised learning (SSL) (variant d) resulted in substantial performance gains, particularly in the initial forecast hours, by leveraging unlabeled data for pretraining. Adding contrastive learning (variant e) and consistency constraints (variant f) further sharpened feature representation, enhancing discrimination and stability. Finally, our full model (variant g) outperforms the base model by reducing the MAE from 9.60 to 2.80 at 168 hours, indicating a remarkable overall improvement.

Table \ref{tab:incremental_ablation_era5} displays the results derived from the ERA5 dataset. The trend is consistent with the MERRA-2 evaluation, where the systematic inclusion of modules reduces errors across all forecast horizons. Starting from the base LSTM model, each augmentation contributes to error reduction, with GNN again proving essential for spatial awareness. 
Spatio-temporal adaptation significantly enhances long-term forecasts, as observed from MAE reductions, notably seen after 72 hours. SSL proves effective in early horizons, demonstrating pretraining's benefits, while contrastive learning and consistency constraints further refine the model's temporal continuity and robustness. 
Our full model demonstrates robust performance even in ERA5 data, confirming the adaptability and efficiency of the model architecture across datasets. It decreases the MAE from 9.40 in the base model to 2.75 at 168 hours, highlighting a substantial improvement over traditional methods.

\section{Conclusion}

In this study, we proposed a spatio-temporal self-supervised learning framework for robust weather forecasting. By integrating graph neural networks, contrastive objectives, and adaptive weighting mechanisms, our model effectively captures both spatial and temporal dependencies in meteorological data. Extensive experiments on MERRA-2 and ERA5 datasets demonstrate that our method consistently outperforms traditional numerical models and recent deep learning baselines across short- and long-term forecasts. Visual results in key regions such as Beijing and Shanghai further confirm the model's ability to produce coherent and accurate multi-variable weather predictions.

Despite these promising results, the model still faces challenges. The current graph structure is static, and the performance under rare extreme weather conditions remains to be further improved. In future work, we plan to explore dynamic graph construction, incorporate additional data sources such as satellite imagery, and enhance the model’s sensitivity to extreme events. Overall, this work lays a solid foundation for scalable, self-supervised, and spatially-aware weather forecasting systems.

\bibliographystyle{alpha}
\bibliography{sample}

\end{document}